# MolCLIP: A Molecular-Auxiliary CLIP Framework for Identifying Drug Mechanism of Action Based on Time-Lapsed Mitochondrial Images

Fengqian Pang[1], Chunyue Lei[1], Hongfei Zhao[2], Chenghao Liu[3], Zhiqiang Xing[1], Huafeng Wang[1], and Chuyang Ye[3(✉)]

[1] North China University of Technology
`fqpang@ncut.edu.cn`
[2] Beijing Neusoft Medical Equipment CO., Ltd
`zhaohongfei@neusoftmedical.com`
[3] Beijing Institute of Technology
`chuyang.ye@bit.edu.cn`

**Abstract.** Drug Mechanism of Action (MoA) mainly investigates how drug molecules interact with cells, which is crucial for drug discovery and clinical application. Recently, deep learning models have been used to recognize MoA by relying on high-content and fluorescence images of cells exposed to various drugs. However, these methods focus on spatial characteristics while overlooking the temporal dynamics of live cells. Time-lapse imaging is more suitable for observing the cell response to drugs. Additionally, drug molecules can trigger cellular dynamic variations related to specific MoA. This indicates that the drug molecule modality may complement the image counterpart. This paper proposes MolCLIP, the first visual language model to combine microscopic cell video- and molecule-modalities. MolCLIP designs a molecule-auxiliary CLIP framework to guide video features in learning the distribution of the molecular latent space. Furthermore, we integrate a metric learning strategy with MolCLIP to optimize the aggregation of video features. Experimental results on the MitoDataset demonstrate that MolCLIP achieves improvements of 51.2% and 20.5% in mAP for drug identification and MoA recognition, respectively.

## 1 Introduction

Drug mechanism of action (MoA) explains how a drug interacts with its target of an organism. Understanding the MoA of small molecule drugs is crucial for guiding the selection, optimization, and clinical development of lead compounds. However, traditional methods for MoA identification generally require various experimental techniques to assess a drug comprehensively [1-3]. High-throughput technologies enable researchers to investigate MoA by mining vast data. For instance, Carraro et al. utilized RNA-seq to analyze gene expression in two cancer cell lines, revealing the MoA at the epigenetic level [4]. A new perspective was proposed to employ proteomics to analyze the MoA of small molecule drugs [5]. Schuhknecht et al. conducted experiments and collected metabolic results using thousands of drugs administered to lung cancer cells, thereby providing in-depth insights into the mechanism of action from a metabolomics standpoint [6].

With the development of computer vision, MoA identification based on cellular phenotypes has emerged. Janssens et al. proposed an unsupervised deep learning algorithm to analyze high-content images of cells treated with various compounds [7]. This algorithm extracts cellular features through a deep network and clusters the features to recognize MoAs. Its superior performance on the BBBC021 dataset demonstrates the feasibility of using cellular phenotypes to explore MoA. Additionally, Seal et al. predicted mitochondrial toxicity by combining cell fluorescence images, gene expression, and compound Morgan fingerprints [8]. Zhang et al. introduced an end-to-end deep neural network to investigate MoA based on the single-cell fluorescence images [9]. Compared to traditional methods, deep learning-based models significantly improved performance. In [10], Tian et al. fused Morgan fingerprints of drugs and cell fluorescence images by multilayer perceptron.

Nevertheless, these studies primarily rely on high-content or fluorescence static images of cells under drug treatment, thereby overlooking cellular dynamics over time. Drug-cell interactions are a time-sustained process. Compared to static microscopic images, time-lapse videos can provide more comprehensive dynamic information. Furthermore, Yu et al. suggested that observing drug-to-cell effects with time-lapse microscopic cell sequences offers valuable insights [11]. This approach successfully identified and confirmed the key role of the natural compound epicatechin in tea on inhibiting cyclooxygenase-2.

MoA describes the interactive approach between cells and drugs, implying significance of drug information. However, previous studies have not deeply utilized drug information. To address this issue, we propose a multi-modality framework, MolCLIP, which incorporates drug information and time-lapse cell sequences. Specifically, it firstly employs two encoders to extract high-dimensional features from drug molecules and cell image sequences. Secondly, an improved CLIP is designed to align the features from these two modalities. The motivation of the designed CLIP stems from the complexity of MoA, which includes coexisting video-drug and video-MoA relationships. Thus, it is essential to enable category awareness in the symmetric supervision matrix of CLIP. Additionally, a metric learning strategy is employed to further constrain cell features, enhancing its recognition capability.

The contributions of this paper are summarized as follows: 1) MolCLIP is the first vision-language model that integrates cell time-lapse sequences with drugs. 2) We proposed a self-awareness and class-awareness supervision matrix designed to optimize the attention mechanism of CLIP for MoA-related applications. 3) Experimental results show that MolCLIP significantly improves the mAP by 51.2% for drug recognition and by 20.5% for MoA indetification. This demonstrates that MolCLIP successfully incorporates drug information into the cell feature space, aiding in the recognition of drug mechanisms.

## 2 Method

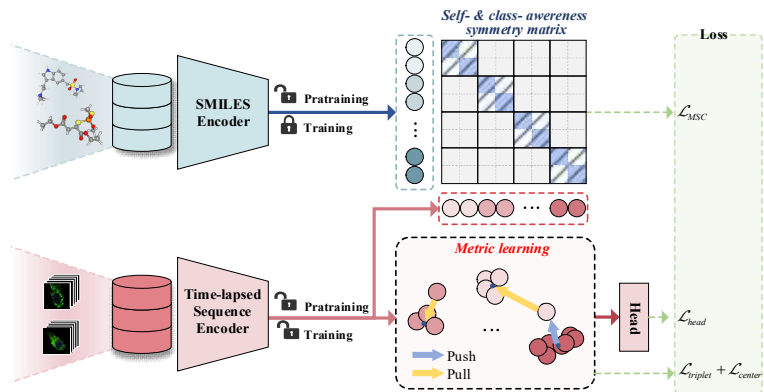

**Fig. 1.** Overview of the proposed MolCLIP framework.

In this section, we first introduce the overview of the MolCLIP. Then, a detailed explanation of encoders is described. Finally, we provides the dual branch of the MolCLIP.

### 2.1 Overview

The MolCLIP framework is depicted in Fig. 1. Assume there are $B$ pairs of samples $\{\mathbb{S}_i, \mathbb{V}_i, l_i^{drug}, l_i^{MoA}\}_{i=1}^{B}$, where $\mathbb{S}_i$ represents the SMILES of the $i-th$ drug. $\mathbb{V}_i$ therein denotes the corresponding time-lapsed sequence. $l_i^{drug}$ and $l_i^{MoA}$ denote the drug type and MoA labels, respectively. $\mathbb{S}_i$ and $\mathbb{V}_i$ are fed into SMILES and Time-lapsed Sequences Encoders ($E_S(\cdot)$ and $E_V(\cdot)$) to obtain high-dimensional features, $S_i$ and $V_i$. $S_i$ and $V_i$ are then aligned via self-awareness and category-aware symmetric matrixes (top right corner of Fig. 1). Meanwhile, the cell feature $V_i$ employs metric learning to enhance classification ability (bottom right corner of Fig. 1). During inference, only the sequence encoder is used to extract dynamic phenotypic features of cells for MoA identification, reducing computational costs.

### 2.2 Encoders

**SMILES Encoder.** MolFormer, a large chemical language model, is used as the SMILES encoder for MolCLIP, which captures the structure and properties of molecules. It first converts SMILES $\mathbb{S}$ into a fixed-length token sequence using a tokenizer with a vocabulary of 2,362 tokens. The token sequence is then fed into a Transformer-based encoder to extract the features of molecules $S \in \mathbb{R}^{1 \times 2048}$.

**Time-lapsed Sequences Encoder.** MolCLIP utilizes I3D ResNet-50 to extract cellular phenotypic variation from time-lapse sequences. Each sequence $\mathbb{V} \in \mathbb{R}^{T \times H \times W \times 3}$ contains $T$ RGB images with a default height $H$ and width $H$ of 112. The time-lapsed sequences encoder $E_V(\cdot)$ extract high-dimensional phenotypic features $V \in \mathbb{R}^{1 \times 2048}$ from the input sequences $\mathbb{V}$, and a linear layer serves as the classification head.

### 2.3  Dual Branch

**CLIP Branch.** Inspired by CLIP [12] framework, we adopt it to align drug information and cellular phenotypic changes to improve the MoA recognition. However, there is a problem when applying CLIP to the MoA recognition task. Unlike the traditional pattern where each image corresponds to a unique description, a drug may correspond to multiple videos, and different drugs may belong to the same MoA. In this case, directly using the symmetric matrix from CLIP would lead to semantic confusion during modality alignment. Therefore, we construct a Self- and Class-awareness symmetry Matrix (SCM) to facilitate MoA learning. In detail, SCM inherits the cross-modal correlation matrix of CLIP and supervises the same-category positions. For this purpose, we design a self-awareness supervision matrix $\mathbf{M}_{self}$ and a class-awareness supervision matrix $\mathbf{M}_{class}$ as follows:

$$\mathbf{M}_{self} = \mathbf{I}_B; \mathbf{M}_{class} = \left[ a_{m,n} \right]_{B \times B}, a_{m,n} = \begin{cases} 1 & s.t. \ l_m^{MoA} = l_n^{MoA} \\ 0 & s.t. \ l_m^{MoA} \neq l_n^{MoA} \end{cases} \quad (1)$$

Moreover, we obtain the similarity matrix $\mathbf{SV}$ based on the features of drugs and their corresponding time-lapsed sequences, which is written as:

$$\mathbf{SV} = \begin{bmatrix} S_1 & S_2 & \cdots & S_B \end{bmatrix}^T \bullet \begin{bmatrix} V_1^T & V_2^T & \cdots & V_B^T \end{bmatrix} \in \mathbb{R}^{B \times B} \quad (2)$$

We employ cross entropy function $\mathrm{CE}(\bullet)$ to form the loss function of CLIP branch, which can be expressed as:

$$\mathcal{L}_{MSC} = \mathrm{CE}(\mathbf{SV}, \mathbf{M}_{self}) + \mathrm{CE}(\mathbf{SV}, \mathbf{M}_{class}) \quad (3)$$

**Metric Learning Branch.** MolCLIP utilizes metric learning strategies to constrain the features of time-lapse sequences, enhancing the classification ability of phenotypic features. Specifically, we first employ the time-lapse sequences encoder to extract phenotypic features of cells, and then we supervise these features using hard triplet loss and center loss. The hard triplet loss can cluster similar feature samples and push dissimilar samples apart. Suppose $V_A$ is the anchor sample, while $H_P$ and $H_N$ denote the hardest positive and negative samples, respectively. The hard triplet loss is formulated as follows:

$$\mathcal{L}_{triplet} = \max\{0, \|V_A - H_P\|_2^2 - \|V_A - H_N\|_2^2 + \lambda_{marg}\} \quad (4)$$

where $\lambda_{marg}$ is a margin threshold hyperparameter and set to 0.3 by default. Furthermore, we use center loss to constrain the time-lapse features. We denote $C_{GT_i}$ as the feature center of the time-lapse sequence's category $GT_i$, and the center loss can be defined as:

$$\mathcal{L}_{center} = \frac{1}{2}\sum_{i=1}^{B}\left\|V_i - C_{GT_i}\right\|^2 \tag{5}$$

Finally, the optimized features are used for MoA recognition through a linear layer, and we supervise the predictions with cross-entropy loss.

## 3 Experiments

### 3.1 Dataset and Preprocessing

**Dataset.** We verified the effectiveness of the proposed MolCLIP using the MitoDataset, which was released by Yu et al. from Zhejiang University [11]. This dataset comprises 35,631 time-lapse sequences under 1,068 FDA-approved drugs and dimethyl sulfoxide (DMSO) treatments. Each sequence consists of 16 frames, totaling 570,096 single-cell images. The Mechanism of Action (MoA) labels for this data are sourced from the Drug Repurposing Hub [13], ChEMBL [14], and DrugBank [15] databases. For the study, they retained 38 drug MoAs and randomly divided them into a training set and a test set in a ratio of 8:2. In the test set, one sample from each drug MoA is randomly selected as the query set, while the remaining samples serve as the gallery set.
**Preprocess.** Since the MitoDataset lacks the SMILES of drugs, we collected them based on MoA label sources, including the Repurposing Hub, ChEMBL, and DrugBank databases. After collecting the data, we used the RDKit library [16] to normalize the SMILES.

### 3.2 Implementation Details

**Implementation details.** All experiments are conducted on an Nvidia 3090 GPU. The time-lapsed sequence encoder employs an improved I3D ResNet-50 architecture, while the SMILES encoder uses the MolFormer XL (MolFormer XL: https://github.com/IBM/molformer). For both pre-training and training, we run 500 epochs with a batch size of 64 by default. SGD is employed as an optimizer with an initial learning rate of 0.001. We first pre-train MolCLIP with drug labels without freezing any network parameters. Subsequently, we fine-tune the model using MoA labels and freeze the SMILES encoder so that it does not participate in the training.

**Metrics.** This paper employs the Cumulative Matching Characteristic (CMC) curve, accuracy, and mean Average Precision (mAP) to evaluate the effectiveness of model.

**Table 1.** Performance of drug recognition under different pretraining strategies (%). The "Seq" in the table indicates that the model relies on cell time-lapse sequences. In contrast, "Seq+Drug" signifies that the model incorporates both cell time-lapse sequences and drug information for drug recognition.

|    | Seq | Seq + Drug | Rank-1 ↑ | mAP ↑ | Acc ↑ |
|----|-----|------------|----------|-------|-------|
| S1 | ✔   | -          | 33.64    | 29.11 | 75.22 |
| S2 | -   | ✔          | 40.20    | 37.20 | 83.67 |
| S3 | ✔   | ✔          | 89.50    | 80.30 | 84.00 |

### 3.3  Exploration of Pretraining Strategy

This section focuses on investigating whether utilizing drug information to assist cell sequences can effectively address the issue of drug recognition. To this end, we constructed a series of experiments aimed at comparing the drug recognition performance of the model. In detail, the comparison experiment was conducted with and without the inclusion of drug information, and the results are summarized in Table 1. The "Seq" in the table indicates that the model only uses cell time-lapse sequences, while "Seq+Drug" expresses that the model uses both cell time-lapse sequences and drug information for drug recognition.

Table 1 contains three drug recognition training strategies: 1) Training the sequence encoder for drug recognition without utilizing drug information. 2) Using the MolCLIP framework to combine drug and cell phenotype information, with the sequence encoder initialized by Kinetic-400 weights. 3) Inheriting the second strategy but initializing the sequence encoder with drug pretrain weight (it can obtain in the first strategy).

It can be observed that the second strategy outperforms the first by 6.56%, 8.09%, and 8.45% in Rank-1, mAP, and Accuracy, respectively. The difference between these two strategies is that the second one introduces drug clues. Therefore, we can infer that performance improvement stems from modeling the drug information.

Subsequently, we explore whether MolCLIP requires a well-initialized sequence encoder (the third strategy). Compared to the second and third rows of Table 1, it shows that the Rank-1, mAP, and accuracy respectively improve by 40.3%, 43.1%, and 0.33% when the sequence encoder initialized by the drug pertaining weights. The only difference between the second and third strategies is whether the sequence encoder is initialized with drug-specific weights.

As aforementioned we employed the metric learning strategy for the cell sequence encoder to enhance the recognition ability of cell features. From this perspective, the MolCLIP can be divided into two parts: the modality alignment part and the metric learning part. Reconsidering the third and second rows, the third one involves more training for the metric learning part from the loss function perspective. Therefore, we increase the loss weight for this part, hoping to achieve comparable performance to the third strategy without additional pre-training. The experiments showed that as the intensity of the metric learning part increased, network performance improved a lot compared to the second strategy. And especially for the accuracy metric exceeded that of the third strategy. However, the poor Rank-1 indicates that the clustering degree of similar features of cells was insufficient. To address the issue, we revisited our previously explored strategies.

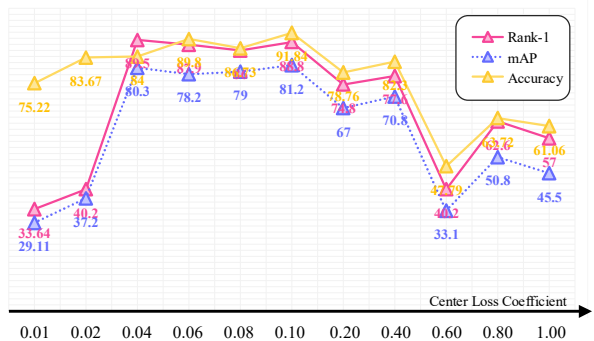

**Fig. 2.** Impact of Center Loss proportion on drug recognition performance (%). The x-axis represents the proportion of the center loss. The y-axis shows the Rank-1, mAP, and Accuracy metrics achieved by the network under each proportion.

In the metric learning part, the performance of MolCLIP is mainly driven by the cross-entropy loss, the hard triplet loss, and the center loss. The cross-entropy loss enables the model to classify the cell features, and the improved accuracy means this function works. The hard triplet loss and center loss focus on the relative and absolute distances between features, respectively. However, there is a problem of insufficient feature clustering in the cell feature space. Therefore, the subsequent research will focus on increasing the proportion of center loss to resolve the issue of insufficient feature clustering.

Initially, we set the coefficient of center loss to 0.01 and kept the proportion of the other loss functions fixed. Then, we gradually increased the center loss coefficient from 0.02 to 0.1 with an interval of 0.02 and from 0.1 to 1.0 with an interval of 0.2. The performance of the network was collected in Fig. 2. We observed that the network performance continuously improved as the center loss weight increased from 0.01 to 0.1 and the model attained the highest accuracy when the center loss weight was equal to 0.1. Notably, the drug recognition accuracy reached 91.84%, which was 7.84% higher than the third strategy in Table 1. Although Rank-1 was still lower than the third strategy by 0.7%, this method (merely adjusting the proportion of center loss) reduced the computational burden of the pre-training process. Therefore, we set the default center loss coefficient to 0.1 for the rest of the study, while keeping the coefficients of the other loss functions at 1.

### 3.4   Comparison with Mainstream Methods

We compare our MolCLIP with six baseline models, and the results are summarized in Table 2. The comparison methods are mainly divided into two categories: 1) machine learning-based methods, including Random Forest [17], Support Vector Machine (SVM) [18], and Multi-Layer Perceptron (MLP) [19], and 2) deep learning-based methods, such as Long Short-Term Memory (LSTM) [20], MitoReID [11], and CLIP [12].

Table 2. MoA identification performance of different methods under MitoDataset (%).

| Methods | Accuracy↑ | Rank-1↑ | Rank-5↑ | Rank-10↑ | mAP↑ |
| --- | --- | --- | --- | --- | --- |
| Random Forest | 15.72 | \ | \ | \ | \ |
| SVM | 10.76 | \ | \ | \ | \ |
| MLP | 12.65 | \ | \ | \ | \ |
| LSTM | 12.18 | \ | \ | \ | \ |
| MitoReID | 82.46 | 76.32 | 84.21 | 84.21 | 65.92 |
| CLIP | 69.91 | 0.00 | 2.80 | 3.70 | 1.10 |
| **Ours** | **93.86** | **91.20** | **94.70** | **94.70** | **86.40** |

These machine learning models classify MoA based on handcrafted features of cell images, including mitochondrial membrane potential density and its changes, as well as mitochondrial area and aspect ratio. The results are shown in the first three rows of Table 2. Experiments demonstrate that these methods exhibit low accuracy and are insufficient for identifying drug MoA. It is speculated that this is due to the complexity of drug mechanisms, making it difficult to accurately classify them solely based on simple cell morphological features.

MitoReID recognizes MoA by solely relying on cell dynamic sequences and achieves 82.46% accuracy, as shown in the fifth row of Table 2. The sixth row displays the performance of CLIP. Although CLIP integrates cell phenotype and drug features, its final accuracy is only 69.91%. This lower accuracy is speculated to CLIP only aligning the two modalities without further constraining the relationships between different drugs of the same mechanism. In fact, the Rank-1 of CLIP in the experiment is 0.0%, which confirms the lower feature clustering in the cell feature space.

MolCLIP utilizes the drug modality auxiliary framework to identify MoA through cell sequences. It introduces drug information to cell feature space by aligning the cell sequence and drug modality. Additionally, it employs metric learning to constrain sequence features, ultimately achieving an accuracy of 93.86%. Furthermore, it attains a Rank-1 of 91.2%, which is an 14.88% improvement over the scenario without introducing drug information. This indicates that MolCLIP successfully incorporates drug features, and these features are beneficial to MoA identification.

## 4    Conclusion

This paper focuses on identifying drug mechanisms of action (MoA) based on cell phenotypes and proposes a novel multi-modalities framework, MolCLIP. To the best of our knowledge, our proposed model presents the first framework that integrates cell dynamics and SMILES for identifying drug MoA. The method incorporates modality alignment as an auxiliary branch, introducing drug information into the phenotype space. However, the experiment demonstrates that modality alignment alone cannot effectively constrain cell dynamic representation, so we employ metric learning to further constrain phenotype features in high-dimensional space. Ultimately, the proposed method significantly improves performance compared to methods without drug information. This indicates that MolCLIP successfully integrates drug information and demonstrates that the drug feature is helpful in recognizing MoA.

**Disclosure of Interests.** The authors have no competing interests to declare that are relevant to the content of this article.

# References

1. Zheng C C, Liao L, Liu Y P, et al.: Blockade of Nuclear β‑Catenin Signaling via Direct Targeting of RanBP3 with NU2058 Induces Cell Senescence to Suppress Colorectal Tumorigenesis. Advanced Science **9**(34), 2202528 (2022)
2. Hu X, Gan L, Tang Z, et al.: A Natural Small Molecule Mitigates Kidney Fibrosis by Targeting Cdc42‑mediated GSK‑3β/β‑catenin Signaling. Advanced Science **11**(13), 2307850 (2024)
3. Wang P, Huang B, Liu Y, et al.: Corynoline protects chronic pancreatitis via binding to PSMA2 and alleviating pancreatic fibrosis. Journal of Gastroenterology **59**(11), 1037-1051 (2024)
4. Carraro C, Bonaguro L, Schulte-Schrepping J, et al.: Decoding mechanism of action and sensitivity to drug candidates from integrated transcriptome and chromatin state. Elife 11, 78012 (2022)
5. Mitchell D C, Kuljanin M, Li J, et al.: A proteome-wide atlas of drug mechanism of action. Nature Biotechnology **41**(6), 845-857 (2023)
6. Schuhknecht L, Ortmayr K, Jänes J, et al.: A human metabolic map of pharmacological perturbations reveals drug modes of action. Nature Biotechnology, 1-13 (2025)
7. Janssens R, Zhang X, Kauffmann A, et al.: Fully unsupervised deep mode of action learning for phenotyping high-content cellular images. Bioinformatics **37**(23), 4548-4555 (2021)
8. Seal S, Carreras-Puigvert J, Trapotsi M A, et al.: Integrating cell morphology with gene expression and chemical structure to aid mitochondrial toxicity detection. Communications Biology **5**(1), 858 (2022)
9. Zhang Z, Chen H, Huang S.: Detection of missing insulator caps based on machine learning and morphological detection. Sensors **23**(3), 1557 (2023)
10. Tian G, Harrison P J, Sreenivasan A P, et al.: Combining molecular and cell painting image data for mechanism of action prediction. Artificial Intelligence in the Life Sciences 3, 100060 (2023)
11. Yu M, Li W, Yu Y, et al. Deep learning large-scale drug discovery and repurposing. Nature Computational Science 4(8), 600-614 (2024)
12. Radford A, Kim J W, Hallacy C, et al.: Learning transferable visual models from natural language supervision. In: International conference on machine learning. pp. 8748-8763. PmLR (2021)
13. Corsello S M, Bittker J A, Liu Z, et al.: The Drug Repurposing Hub: a next-generation drug library and information resource. Nature medicine 23(4), 405-408 (2017)

14. ChEMBL Homepage, https://www.ebi.ac.uk/, last accessed 2025/02/27
15. DrugBank Homepage, https://www.drugbank.com/datasets, last accessed 2025/02/27
16. RDKit Homepage, https://www.rdkit.org/, last accessed 2025/02/27
17. Breiman L.: Random forests. Machine learning 45, 5-32 (2001)
18. Hearst M A, Dumais S T, Osuna E, et al.: Support vector machines. IEEE Intelligent Systems and their applications **13**(4), 18-28 (1998)
19. Popescu M C, Balas V E, Perescu-Popescu L, et al.: Multilayer perceptron and neural networks. WSEAS Transactions on Circuits and Systems 8(7), 579-588 (2009)
20. Hochreiter S, Schmidhuber J.: Long short-term memory. Neural computation 9(8), 1735-1780 (1997)